# TOWARDS A COMMON SPEECH ANALYSIS ENGINE

*Hagai Aronowitz*[1], *Itai Gat*[1,2], *Edmilson Morais*[1], *Weizhong Zhu*[1], *Ron Hoory*[1]

[1]IBM Research AI
[2]Technion - Israel Institute of Technology

## ABSTRACT

Recent innovations in self-supervised representation learning have led to remarkable advances in natural language processing. That said, in the speech processing domain, self-supervised representation learning-based systems are not yet considered state-of-the-art.

We propose leveraging recent advances in self-supervised-based speech processing to create a common speech analysis engine. Such an engine should be able to handle multiple speech processing tasks, using a single architecture, to obtain state-of-the-art accuracy. The engine must also enable support for new tasks with small training datasets. Beyond that, a common engine should be capable of supporting distributed training with client in-house private data.

We present the architecture for a common speech analysis engine based on the HuBERT self-supervised speech representation. Based on experiments, we report our results for language identification and emotion recognition on the standard evaluations NIST-LRE 07 and IEMOCAP. Our results surpass the state-of-the-art performance reported so far on these tasks.

We also analyzed our engine on the emotion recognition task using reduced amounts of training data and show how to achieve improved results.

## 1. INTRODUCTION

Over the past years, advances in speech processing have been remarkable, due to deep learning-based end-to-end systems and to the growing availability of large datasets for training.

However, many challenges remain. First, for many tasks the available training data is not large enough to optimally train deep models. Second, the need to independently train a separate system for every task is inefficient both in terms of data usage and in terms of research, development, and maintenance processes. Third, many clients have in-house datasets which are typically both small and private.

Self-supervised representation learning can play a significant role in tackling the above challenges, due to its ability to learn strong feature representations from unlabeled task-independent data. Recently, this framework has shown success in the speech processing community.

We use the recently introduced HuBERT [1] framework to extract speech representations. Next, we use simple temporal average pooling and a linear layer on top of the HuBERT extractor to carry out classification. Evaluations show that using the proposed approach we outperform the currently published state-of-the-art for both language identification and emotion recognition. We describe several simple techniques that have a significant impact on accuracy. Furthermore, we analyze the performance of the proposed architecture when training data is very limited.

The remainder of the work is organized as follows: Section 2 provides an overview of related work. Section 3 describes our proposed common speech analysis engine. Section 4 reports the language identification experiments and results, and Section 5 reports the emotion recognition experiments and results. Finally, Section 6 concludes the paper.

## 2. RELATED WORK

### 2.1. Self-supervised speech representation-based speech analysis

In natural language processing (NLP), self-supervision has proven to be an extremely successful approach for training representation extractors [2]. Self-supervision has been used to leverage huge amounts of unlabeled data to train a language model using masked prediction. In this framework, transformers [2] have been very successfully in modeling long contexts. The high quality of the resulting representations enables the training of simple and powerful task dependent classifiers on top of the representations.

The process of extracting task-independent representations is referred to as *upstream* processing. The subsequent process of task-dependent classification is known as *downstream* processing.

Following the success of self-supervision in NLP, researchers in the speech community started to explore the use of self-supervision in the speech domain. That said, there are some major challenges in adapting self-supervised techniques from NLP to the speech domain. First, speech signals are continuous-valued sequences, which contrasts with the discrete nature of words in NLP and hinders the use of predictive losses. Second, the boundaries between sound units are not known, which complicates masked prediction pre-training.

During the past two years, several self-supervised frameworks for extracting speech representations were introduced, including wav2vec [3], TERA [4], TRILL [5], wav2vec2.0 [6] and HuBERT [1]. The quality of these upstream models is constantly improving due to improved modeling and the availability of larger unlabeled training dataset.

In our initial experiments we observed that the publicly available models for HuBERT and wac2vec 2.0 are significantly superior to those available for wav2vec, TERA and TRILL. The rest of our experiments are mostly based on the HuBERT framework.

Hidden-Unit BERT (HuBERT) for self-supervised speech representation learning is based on masking regions of the audio signal, quantizing those regions using a clustering mechanism, and predicting the discrete clustering assignments of the masked regions using the unmasked regions. The predictive loss forces the model to learn good high-level representations of unmasked inputs to infer the targets of the masked ones. The HuBERT model learns both acoustic and language models from continuous inputs by modeling unmasked inputs into meaningful continuous latent representations and capturing the long-range temporal relations between learned representations.

## 2.2. Spoken language identification

Spoken language identification (LID) is the task of identifying the spoken language in a speech utterance. Typically, LID systems are composed of a frame-level feature extraction step followed by an utterance-level feature extraction step that aggregates frame-level information, and finally a classification step.

NIST language recognition evaluations (LRE) have been the major LID evaluation datasets for the past 25 years [7]. The NIST 2007 LRE [8] is the most popular one in published literature and is still used significantly in current publications. We therefore use it to evaluate our LID work in this paper. The NIST 2007 LRE defines an LID task for the classification of 14 languages and defines training, development, and test sets, although many recent works use additional training data. Similar to other works, we report our results in equal error rate (EER) on the 14-languages closed-set task, on 3s, 10s and 30s test durations.

Shen et al. [9] feed log mel-filterbank features to an x-vector extractor to first extract frame-level features using a time delay neural network (TDNN); they then pool the temporal mean and standard deviations and further process them to produce an utterance level embedding that is then classified into language classes. The results reported are presented in 1.

Cai et al. [10] feed speech represented by a sequence of filter-banks to a ResNet-34-based end-to-end network. This is followed by a learnable dictionary encoding (LDE) layer which pools the temporal features into an utterance-based embedding. The pooling and embedding is done by to a mechanism that works similar to the way a GMM supervector embeds speech utterances [11]. Finally, the utterance-based embedding is fed to a classification layer. The results reported are presented in Table 1.

In [12], the authors trained a bottleneck feature extractor based on a multilingual conformer by jointly training a multilingual Automatic Speech Recognition (ASR) system and an LID system, both sharing the same feature extractor. The LID system was based on a ResNet architecture. This approach produced the best published results so far. The results are presented in Table 1.

Leveraging self-supervised speech representations for LID is something that has scarcely been done. Shor et al. [5] composed a TRILL based representation extractor with average pooling and a linear downstream classifier. The TRILL embedding is finetuned to the LID training data. The results obtained on VoxForge [13] are reported in Table 2.

Ramesh et al. [14] used wav2vec as an upstream model with a convolutional recurrent neural network with attention as a downstream model; they report an accuracy improvement on an evaluation of seven Indian languages.

Tjandra et al. [15] used wav2vec 2.0 trained on the multilingual XLSR dataset [16] to identify 25 languages. The downstream model is a simple average pooling followed by a linear classifier. The authors did not compare their system to other state-of-the-art approaches.

## 2.3. Speech emotion recognition

Speech emotion recognition (SER) is the task of classifying a speaker's emotional state into a single emotion category. There is a large amount of literature on SER. Typically, SER systems are composed of a feature extraction step followed by a classification step. Today, the best works are based on deep learning, mostly done in an end-to-end fashion.

IEMOCAP [17] is the most widely evaluated dataset in the SER literature. IEMOCAP contains approximately 12 hours spoken by 10 actors, divided into 5 couples. The conversation between each couple constitutes a session and is split into utterances. The data consists of several thousands of utterances, each labeled with a single emotion category. The emotion categories are standardly merged into 4 categories: anger, happiness, sadness, and neutrality. The organization of IEMOCAP into train, development and test sets has not been officially defined, and different works define these sets differently. The most common setup is to use 5-fold cross validation, i.e., for each fold, 8 speakers are used for training and 2 for testing. The corresponding development set is extracted from the training set. We denote this setup as *5-fold*.

Some works carry out 5-fold cross-validation by randomly dividing the utterances into train and test sets without making sure the speakers in the train and test sets are disjoint. One recent work referred to this setup as speaker-dependent [18]. In this setup, the measured accuracy is expected to be higher. Note that IEMOCAP consists of both improvised speech (easier) and scripted speech (harder); some works evaluate only on the easier improvised part. Other works report results only on session no. 5 [19]. We do not consider these works when we review the state-of-the-art because their results are not comparable. We report results using weighted average (WA) [20] in accuracy.

Most published SER works operate only on the audio. Other works operate on both audio and text. The text may be either a manual transcription, or an automatic transcription of the audio. The accuracy obtained using this multi-modal approach is obviously higher: 76.4% using manual transcription [21] and 68.6% using automatic speech recognition [22]. Our work focuses on the audio modality without using automatic transcription resources.

The following is a review of the current state-of-the-art. Li et al. [23] report an accuracy of 71.8% by feeding a spectrogram into a CNN with time-specific and frequency-specific filters, followed by attention-based pooling. Wang et al. [24] applied CNN layers followed by a dual sequence LSTM to obtain an accuracy of 69.4 which goes up to 72.7 when fused with another system. Zhao et al. [25] developed an attention-based bidirectional long short-term memory (BLSTM) neural network in combination with a connectionist temporal classification (CTC) objective function (Attention-BLSTM-CTC) and obtained an accuracy of 69%.

Recently, self-supervised speech representations have been successfully used for SER. Pepino et al. [26] obtained an accuracy of 67.2% using wav2vec 2.0 as an upstream model (with finetuning). Their downstream model is composed of 2 dense layers followed by temporal average pooling followed by another dense layer.

Yang et al., [27] report an accuracy of 67.6% using HuBERT as an upstream model (without finetuning) and temporal average pooling with a simple linear layer as a downstream model.

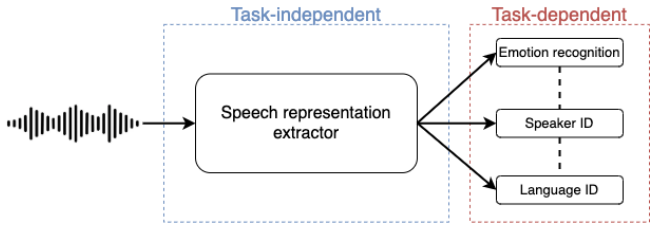

Figure 1: Proposed architecture for the common speech analysis engine. The upstream processing is illustrated in blue, and the downstream processing is illustrated in red.

X. Cai et al. [20] report an accuracy of 78.2% using 10-fold cross validation. This setup is argued in [28] to be biased, especially when the textual modality is explicitly modeled, due to lexical within-session correlation. They use wav2vec 2.0 as an upstream model, and train both an SER downstream model and an automatic speech recognition downstream model in a mutli-task training setup.

Xia et al. [29] obtain an accuracy of 66.9% using a wav2vec upstream model followed by a downstream model composed of 4 CNN layers, a segment layer classifier, a 5-layer BLSTM, average pooling and finally a linear layer.

### 3. COMMON SPEECH ANALYSIS ENGINE

Our proposed architecture for a common speech analysis engine is described in Figure 1. The audio is fed to upstream processing which extracts frame-level speech representations (such as HuBERT). The frame-level representations are fed into downstream processing. The downstream processing first applies temporal average pooling to the representations resulting in a single average representation per audio session. It then applies a classification layer which is a simple linear layer followed by softmax.

#### 3.1. Supporting a new task

Enabling the engine to support new tasks requires the availability of task-dependent labeled training data. There are two options for training the system. The first one is to freeze the upstream model, and only train the linear classification layer. The second option is to jointly finetune the upstream model and train the classification layer. We report results for both approaches.

#### 3.2. Weight averaging

When the available training and development data is relatively limited, as is the typical in SER, the trained models tend to be noisy. Moreover, the stopping criterion becomes challenging, because the correlation between the development accuracy and test accuracy is not perfect.

To obtain a robust model, we propose using stochastic weight averaging [30] which averages several trained models into one by computing the average in weight space. During training, instead of selecting the top performing model for the development data, we select several top-performing models from different epochs. The top-performing models are then averaged in the model's parameter space and the resulting averaged model is used for testing. Thus, no extra cost in terms of resources is needed for the testing phase.

Table 1. *LID EER results for NIST-2007 LRE (in %)*

| System description | 3s | 10s | 30s |
|---|---|---|---|
| x-vectors [9] | 6.9 | 3.0 | 1.6 |
| ResNet34 LDE [10] | 7.8 | 2.3 | 1.0 |
| Joint LID & ASR [12] | 5.8 | 1.5 | 0.7 |
| Common Speech Analysis Engine | | | |
| HuBERT Base frozen | 15.1 | 6.4 | 3.5 |
| HuBERT Large frozen | 12.6 | 3.2 | 2.4 |
| HuBERT Base finetuned | 4.6 | 1.5 | 0.5 |
| HuBERT Large finetuned | **3.8** | **1.3** | **0.4** |

Table 2. *LID accuracy results for VoxForge (in %)*

| System description | Accuracy |
|---|---|
| TRILL [5] | 94.1 |
| x-vectors | 96.1 |
| Common Speech Analysis Engine: HuBERT large finetuned | **99.8** |

### 4. LANGUAGE IDENTIFICATION EXPERIMENTS

We used the s3prl toolkit [31] as a basis for our experiments and modified it to support LID.

#### 4.1. NIST-LRE-07 setup

We used the NIST-LRE-07 dataset for our evaluation. Our training and development datasets consisted of NIST-LREs 03, 05, 07 (train and development), CALLHOME, CALLFRIEND, and NIST-SRE-08 (Part 1). We upsampled all the 8KHz data to 16KHz to match the data to the HuBERT extractor which operates at 16KHz.

We report results on the main task of the 14-language closed set-task NIST-LRE on 3s, 10s and 30s test durations. We report performance in terms of EER.

#### 4.2. VoxForge setup

To compare out results to the TRILL system, we evaluated our solution on VoxForge. The VoxForge LID subset consists of six languages: English, French, German, Spanish, Russian and Italian. These come from the VoxForge [13] dataset, an open-source speech corpus primarily consisting of samples recorded and submitted by users using their own microphone. We follow the organization of [5] to divide the data into training, development, and test sets. Similar to [5], we report performance in terms of accuracy.

#### 4.3. NIST-LRE-07 experiments

Table 1 reports our results on NIST-LRE-07 compared to the state-of-the-art works described in Section 2. We evaluated both the HuBERT-base and HuBERT-large upstream models. The HuBERT-base model was trained on 960 hours of unlabeled read English speech (Librispeech [32]). It consists of 95M parameters and runs 200 times faster than real-time (RT) on a v100 GPU. The HuBERT-large was trained on 60,000 hours of unlabeled audiobooks in English (libri-light [33]). It consists of 316M parameters and runs 70 times faster than real-time.

We explore both using the frozen pretrained HuBERT models and finetuning the HuBERT models on the language ID training data. Our experiments show that finetuning is required to obtain state-of-the-art results. This may be due to the pretrained HuBERT models being trained on English only data. HuBERT-large is significantly better than HuBERT-base when frozen. When finetuned, HuBERT-large reduced EER by 17% compared to HuBERT-base.

The common speech analysis engine with the finetuned large HuBERT upstream model obtains EERs of 3.8%, 1.3% and 0.4% for the 3s, 10s and 30s respectively. This is an average error reduction of 30% relative to the state-of-the-art joint multilingual ASR an LID system [joint-LID&ASR] that is trained on transcribed multilingual speech. The relative error reduction is 52% compared to the state-of-the-art ReseNet34 LDE system [10] which does not require transcribed speech.

### 4.4. VoxForge experiments

Table 2 reports our results on VoxForge compared to the results reported in [5] and to an x-vector-based system we developed as a baseline. The common engine with the finetuned large HuBERT model dramatically outperformed the baseline systems achieving an accuracy of 99.8% compared to 94.1% reported for TRILL and 96.1% when we used an x-vector system. Contrary to the experiments with LRE-07, we did not need upsampling in the VoxForge experiments.

## 5. EMOTION RECOGNITION EXPERIMENTS

For our evaluations and experiments in SER tasks, we used the s3prl toolkit [31] as a basis and modified it to support the enhancements described in Section 3.

### 5.1. IEMOCAP setup

We used the IEMOCAP dataset as described in Section 2. The development set was taken for each fold by extracting 12% of the training data.

### 5.2. Experiments with the full IEMOCAP dataset

Table 3 reports our SER results compared to the state-of-the-art works described in Section 2. We report results using the HuBERT-large upstream model (see Section 4 for details).

We explored using the frozen pretrained HuBERT models and finetuning the HuBERT models on the IEMOCAP training data.

The common speech analysis engine with the finetuned large HuBERT upstream model obtains an accuracy of 73%. This accuracy outperforms the contrasted systems reported in the Table. Averaging the top-5 models according to development set accuracy resulted in a model that obtains an accuracy of 75.2%, which is the current state-of-the-art.

### 5.3. Experiments with reduced amounts of training data

IEMOCAP contains ~1000 training utterances per emotion category. We sampled smaller subsets of training utterances and ran experiments using the reduced subsets. The results are shown in Figure 2.

Table 3. *SER accuracy results for IEMOCAP (in %)*

| System description | Accuracy |
|---|---|
| Wav2vec + CNN + 5-layer BLSTM [29] | 66.9 |
| Wav2vec 2.0 + finetuning [26] | 67.2 |
| HuBERT frozen [27] | 67.6 |
| Attention-BLSTM-CTC [25] | 69.0 |
| Dual sequence LSTM [24] | 69.4 |
| CNN + attention [23] | 71.8 |
| Common Speech Analysis Engine | |
| HuBERT Large frozen | 66.4 |
| HuBERT Large finetuned | 73.0 |
| HuBERT Large finetuned + averaged models | 75.2 |

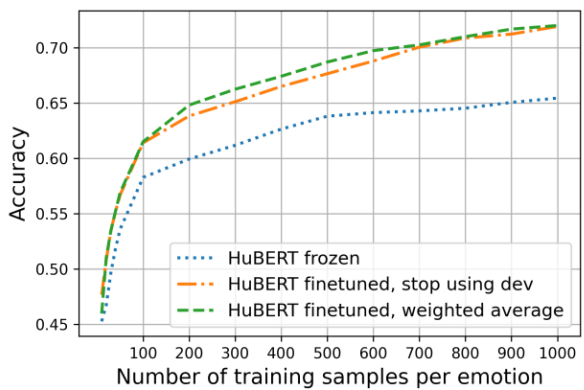

Figure 2: SER accuracy results on reduced training subsets for IEMOCAP (in %)

The small size of the development sets (12% of the corresponding training set) makes it difficult to select the optimal model from the training process, and renders the top-5 weighting scheme less powerful.

Instead, from each training we selected the model that performs best on the development set and the model obtained after 10,000 batches; we then averaged them both in weight space into a single average model.

The model averaging method with finetuned HuBERT-large obtains 65% accuracy with 200 training samples per emotion category, and 70% accuracy with 600 training samples per emotion category. Model averaging is Below 100 training samples training converges very quickly

## 6. CONCLUSIONS

We developed a common speech analysis engine that leverages recent advances in self-supervised-based speech processing. Our experiments show that the engine surpasses the state-of-the-art accuracy for both language identification and emotion recognition on standard evaluations.

For small training data, we show a significant improvement using model averaging in weight space.

The frozen representation can be used for distributed training with private data. Our future plans include improving the common engine results with frozen representations, validating the common engine on additional tasks and handling both speech and text input.